\documentclass[11pt]{article}

\usepackage{dialogue}

\usepackage{ifxetex}
\ifxetex
    \usepackage{fontspec}
    \setromanfont{Times New Roman}
\else
  \usepackage{cmap}
  \usepackage[T2A,T1]{fontenc}
  \usepackage[utf8]{inputenc}
  \usepackage{times}
  \usepackage{latexsym}
  \usepackage{substitutefont}
  \substitutefont{T2A}{\familydefault}{cmr}
\fi

\usepackage[russian,british]{babel}
\usepackage{url}
\usepackage{pgf}

\usepackage{covington} 

\usepackage{hyperref}

\dialogfinalcopy 

\title{Russian Texts Detoxification with Levenshtein Editing}

\author{Ilya Gusev \\
  Moscow Institute of Physics and Technology \\
  Moscow \\
  {\tt ilya.gusev@phystech.edu}}

\date{2022}

\begin{document}
\maketitle
\begin{abstract}
   Text detoxification is a style transfer task of creating neutral versions of toxic texts. In this paper, we use the concept of text editing to build a two-step tagging-based detoxification model using a parallel corpus of Russian texts. With this model, we achieved the best style transfer accuracy among all models in the RUSSE Detox shared task, surpassing larger sequence-to-sequence models.
  
  
  \textbf{DOI:} 10.28995/2075-7182-2022-20-XX-XX
\end{abstract}

\selectlanguage{russian}
\begin{center}
  \russiantitle{Преобразование оскорбительных текстов на русском языке с помощью предсказания редакционных предписаний}

  \medskip \setlength\tabcolsep{2cm}
  \begin{tabular}{cc}
    \textbf{Илья Гусев}\\
      Московский физико-технический институт\\
      Москва\\
      {\tt ilya.gusev@phystech.edu}
  \end{tabular}
  \medskip
\end{center}

\begin{abstract}
  Детоксикация текста --- это задача создания нейтральных версий оскорбительных текстов. В этой статье мы используем концепцию преобразования текста с помощью предсказания редакционных предписаний для построения двухэтапной модели детоксикации русских текстов при наличии параллельного корпуса. С помощью этой модели мы добились наилучшей точности передачи стиля среди всех участаников дорожки RUSSE Detox, превзойдя более крупные sequence-to-sequence модели.
  
  \textbf{Ключевые слова:} оскорбительные тексты, детоксификация, перенос стиля, BERT, T5, тегирование, редакционные предписания
\end{abstract}
\selectlanguage{british}

\section{Introduction}
\label{intro}

There is a vast amount of user-generated content on the Internet containing hate speech, profanity, toxicity, and aggression. It may not be appropriate for some platforms to show toxic texts. Some countries can even consider illegal writing or showing such content.

There are several ways to combat this problem. The obvious solution is to censor all toxic messages. Such texts can be deleted completely, covered with a warning, or placed at the very bottom of the page. However, it is ethically questionable apparent censorship.

Another way is to prevent writing such messages by suggesting alternative neutral options to a user. We will refer to the task of making such neutral variants of toxic texts as detoxification. It is a style transfer task where the source style is toxic, and the target style is neutral. The goal of this work was to build a system to solve this task.

Why is this task difficult?
\begin{enumerate}
    \item Indistinct boundaries of what to consider toxic
    \item Obfuscations that hide the meaning of words
    \item Occasions of some rare insults
    \item Sarcasm and other issues that require external world knowledge
\end{enumerate}

Toxicity is a broad term that includes hate speech, obscene or condescending language, aggression, or grave insults. An instruction for annotators should define the particular rules for it. From the perspective of the shared task organizers, a text should contain <<insults or obscene and rude words>> to be considered toxic.

From a scientific perspective, it is a curious sequence-to-sequence task, where a target text is almost the same as a source one but with a different style. It allows specific methods that rely on the similarity of source and target texts.

This work is a part of the RUSSE Detox shared task, organized by a group of researchers as a part of the Dialogue-2022 conference. The goal of the shared task was to build a detoxification model with provided parallel corpus. Organizers also provided several baselines.

Our contributions:
\begin{enumerate}
    \item We adopt a concept from the LEWIS paper~\cite{lewis} to build a two-step tagging-based detoxification model using a parallel corpus of Russian texts.
    \item We compare this tagging-based model with sequence-to-sequence baselines trained on the same corpus.
    \item We propose a better model for toxicity classification.
    \item We achieve the best style transfer accuracy among all models in the shared task.
\end{enumerate}

Our code\footnote{\url{https://github.com/IlyaGusev/rudetox}} and models\footnote{\url{https://huggingface.co/IlyaGusev/rubertconv_toxic_clf}}\footnote{\url{https://huggingface.co/IlyaGusev/rubertconv_toxic_editor}}\footnote{\url{https://huggingface.co/IlyaGusev/sber_rut5_filler}} are available online.

\section{Related work}

\subsection{Toxicity classification}

\newcite{nobata_abusive_2016} made one of the first attempts to formulate the task of toxicity classification and collect a unified test dataset for it. They used comments posted on Yahoo Finance and News and rated by their in-house workers. The model was Vowpal Wabbit’s regression over different manual NLP features.

\newcite{gordeev} selected anonymous imageboards (4chan.org, 2ch.hk) as the material for their corpus for the task of analysis of aggression. Authors utilized convolutional neural networks to detect the state of aggression in English and Russian texts. 

\newcite{andrusyak} collected a dataset from Russian YouTube comments in an unsupervised way using a seed dictionary of abusive words and an iterative process updating this dictionary.

\newcite{smetanin_toxic} used the Russian Language Toxic Comments Dataset (RTC dataset) from Kaggle\footnote{\url{https://www.kaggle.com/blackmoon/russian-language-toxic-comments}}. It is the collection of annotated comments from 2ch\footnote{\url{https://2ch.hk/}} and Pikabu\footnote{\url{https://pikabu.ru/}} websites. Fine-tuned RuBERT~\cite{rubert} was the best model from this paper.

\newcite{zueva} introduced a novel corpus of 100000 comments posted on a major Russian social network (VK). As their primary model, they used a self-attentive encoder to get interpretable weights for each input token. They also used several tweaks, such as identity dropout and multi-task learning.

\newcite{saitov} utilized Russian subtitles from the <<South Park>> TV show (RSP dataset) and the RTC dataset. Crowdsourcers annotated these subtitles for toxicity. Again, RuBERT held the best result.

\newcite{pronoza} focused on ethnicity-targeted hate speech detection in Russian texts. The authors composed a dataset of 5600 texts with 12000 mentioned instances of different ethnic groups. They named it RuEthnoHate. One more time, the modified RuBERT with additional linguistic features was the best model.

We used several of the mentioned datasets for toxicity classification to fine-tune a conversational RuBERT model.

\subsection{Style transfer}

\newcite{delete_retrieve_generate} started the whole field of research, proposing a set of simple baselines for unsupervised text style transfer. The baselines were based on detecting style tokens with n-gram statistics and replacing them with altered retrieved similar sentences with the target style.

\newcite{condbert} introduced a way to augment texts without breaking the label compatibility. They trained a \textbf{conditional BERT}~\cite{bert} using a conditional MLM task on a labeled dataset. Aside from data augmentation, they proposed to use this method as a part of a style transfer system, using an attention-based method to find style words and conditional BERT to replace them.

\newcite{paraphrase_style_transfer} suggested \textbf{STRAP}, \textbf{S}tyle \textbf{Tra}nsfer via \textbf{P}araphrasing. First, they generated a pseudo-parallel corpus. They started with styled texts and applied paraphrasers to normalize these texts in terms of style. The diversity of paraphrasing was promoted by filtering outputs heavily. Then, they fine-tuned style-changing inverse paraphrasers on this pseudo-parallel corpus. GPT2~\cite{gpt2} language model was used to implement both the paraphrasers and inverse paraphrasers. This scheme can also be used to augment an existing parallel corpus.

They also criticized existing style transfer evaluation methods and proposed an evaluation scheme based on transfer accuracy, semantic similarity, and fluency that we use in this work.

\newcite{masker} introduced \textbf{Masker}, a system that used two language models to detect style tokens and padded masked language models to replace them. They tested it on sentence fusion and sentiment transfer. As for supervised tasks, they created \textbf{LaserTagger}~\cite{laser_tagger}, a sequence tagging approach that casts text generation as a text editing task.

\newcite{gedi} used \textbf{GeDi} (\textbf{Ge}nerative \textbf{Di}scriminator) to control generation towards the desired style. They use three language models: a base one, one for the desired style, and one for the undesired anti-style. The Bayes rule is applied during generation to compute style modifiers for every token from a vocabulary. Then these modifiers are applied to predictions of the base language model. This method allows computationally effective style-guided generation, but there is no source sequence, unlike the style transfer task. \newcite{david_detox} introduced the \textbf{ParaGeDi} method that applies GeDi for style transfer using a paraphrasing model instead of the base language model.

\newcite{skolkovo_detox} introduced the first study of automatic detoxification of Russian texts. They proposed two methods, the unsupervised one based on condBERT and the supervised one based on fine-tuning pretrained language GPT-2 model on a small manually created parallel corpus.

\newcite{lewis} proposed \textbf{LEWIS} (\textbf{L}evenshtein \textbf{E}diting \textbf{WI}th unsupervised \textbf{S}ynthesis), the editing and synthesis framework for text style transfer. They had no parallel data, so the first task was to create a pseudo-parallel corpus. They used an attention-based detector of style words and two style-specific BART~\cite{bart} masked language models to replace these style words. Then they filtered resulting pairs with a style classifier, keeping only examples where the language models and the classifier agree.

After obtaining the pseudo-parallel corpus, they trained a RoBERTa-tagger~\cite{roberta} on it, predicting coarse edit types: <<insert>>, <<keep>>, <<replace>> and <<delete>>~\cite{levenshtein}. Then they trained a fine-grain edit generator to produce the target text, filling in phrases for coarse-grain edit types <<insert>> and <<replace>>. We use this scheme almost without any modifications, but with a different language, with different base models, and already existing parallel corpus.

\section{Evaluation}

We built our style classifier by fine-tuning conversational RuBERT~\cite{rubert} instead of the model\footnote{\url{https://huggingface.co/SkolkovoInstitute/russian\_toxicity\_classifier}} proposed by organizers of the shared task. In addition to ok.ru\footnote{\url{https://www.kaggle.com/datasets/alexandersemiletov/toxic-russian-comments}} and 2ch/Pikabu\footnote{\url{https://www.kaggle.com/datasets/blackmoon/russian-language-toxic-comments}} datasets, we used Russian Persona Chat dataset\footnote{\url{https://toloka.ai/ru/datasets}} as a reliable source of non-toxic sentences.

\begin{table}[t]
\begin{center}
  \begin{tabular}{|c|c|c|c|}
    \hline
    \bf Test type & \bf Test description & \bf Skolkovo clf., ER, \% & \bf Our clf., ER, \% \\
    \hline
    INV & Replace yo & 0.6 & 0.0 \\
    INV & Remove exclamations & 0.9  & 0.4 \\
    INV & Add exclamations & 0.9 & 0.3 \\
    INV & Captioned sentences to lowercase & 73.9 & 34.8 \\
    INV & Remove question marks & 4.0 & 0.2 \\
    INV & Add typos & 3.6 & 1.9 \\
    INV & Masking of characters in toxic words & 5.2 & 0.5 \\
    INV & Add typos to toxic words only & 24.2 & 2.8 \\
    MFT & Concatenate non-toxic and toxic texts & 15.5 & 3.1 \\
    MFT & Concatenate two non-toxic texts & 2.1 & 0.6 \\
    MFT & Add toxic words from a vocabulary & 16.3 & 0.1 \\
    \hline
  \end{tabular}
\end{center}
\caption{Error rates on different tests for two toxic classification models}
\label{tab:checklist} 
\end{table}

We also tested models using a <<checklist>>~\cite{checklist} methodology and augmented the resulting dataset with all the transformations. Test results are in Table~\ref{tab:checklist}. There are invariance (INV) and minimum functionality tests (MFT). Invariance tests ensure that a label will not change after a transformation, and MF tests have a fixed label to be predicted. It is clear from the table that our model has much lower error rates. From the user's perspective, it is harder to pick up an adversarial example for our model than for the default one.

\begin{table}[t]
\begin{center}
\begin{tabular}{|c|c|c|c|}
\hline
\bf Model & \bf AUC, \% & \bf Accuracy, \% & \bf F1, \% \\ \hline
Skolkovo classifier & 66.2 & 86.4 & 37.2 \\
Our classifier & 73.5 & 90.3 & 51.3 \\
\hline
\end{tabular}
\end{center}
\caption{Metrics of toxicity classifiers on unseen crowdsourced test set: 3642 unique texts, 355 of them are toxic}
\label{tab:automatic_clf} 
\end{table}

Two models have different dataset splits, so comparing them on their native test sets is wrong. However, we used crowdsourcing to evaluate the style transfer model, so we can use these annotations as an independent test set, keeping in mind that these samples are adversarial. Results for this new set are in Table~\ref{tab:automatic_clf}. Our classifier shows better results in this setting.

We used models provided by organizers of the shared task for measuring semantic similarity\footnote{\url{https://huggingface.co/cointegrated/LaBSE-en-ru}} and fluency\footnote{\url{https://huggingface.co/SkolkovoInstitute/rubert-base-corruption-detector}}. They have similar problems, but we did not come up with better options.

However, automatic metrics are not reliable, especially when being used with near-adversarial examples. Table~\ref{tab:automatic_clf} gives a glance at how unreliable they can be. To overcome this, we arranged our in-house annotation process with crowdsourcing through the Toloka\footnote{\url{https://toloka.ai}} platform in addition to the final evaluation provided by organizers of the shared task. We measured only style accuracy and semantic similarity, as fluency was much harder to define. Annotation instructions were close as possible to ones provided by the organizers and are available in the repository. Five workers annotated every sample. Samples were aggregated by majority vote. The average agreement was 90\% for the style accuracy project, with Krippendorff’s alpha of 46\%. For the similarity project, the average agreement was 88\%, with Krippendorff’s alpha of 49\%.

\section{Model}
We see text detoxification as a two-step process. In the first step, a model should determine what words should be deleted or replaced. We can explicitly do it through tagging. In the second step, a generator replaces words or adds new ones. From this perspective, any classical sequence-to-sequence model has a trivial first step, as all words can be replaced.

\subsection{Tagger --- first step}
\subsubsection{Based on interpretation of a classifier}
One way to find style tokens is to interpret a classification model. As for attention-based models, one can find such tokens using attention distribution. Tokens with high attention scores correlate with tokens that manifest style. Many researchers used this method~\cite{unpaired_sentiment,condbert,exbert,lewis}.

It is also possible to use models that allow interpretation by design. \newcite{skolkovo_detox} utilized logistic regression and its weights for each word from the vocabulary for this task, and \newcite{delete_retrieve_generate} used a Naive Bayes classifier.

\subsubsection{Based on language models}
Another way is to use two language models, one trained on texts of one style and another trained on texts of a different style. We can calculate the proportion of their predictions for every token if we have such models. If a prediction of the first model is much higher than that of the second model, then a corresponding token can be style-loaded. For instance, Masker~\cite{masker} used a similar approach.

\subsubsection{Based on tags from parallel corpus}
Finally, if we have a parallel corpus, we can directly compute edits required to transform source texts into target texts, convert these edits to tags, and then predict these tags with a token classification model.

\subsection{Generator --- second step}

\subsubsection{Based on MLM models}
One way of filling the gaps is to use models pretrained for masked language modeling tasks (MLM) such as BERT~\cite{bert} or T5~\cite{t5}. It is their original task, but one can fine-tune them on texts of the required style~\cite{condbert}. However, these models have no access to original words, so they can rely solely on context remained after masking.

\subsubsection{Based on pairs from parallel corpus}

\begin{figure}[t]
  \centering
  \includegraphics[width=1.0\linewidth]{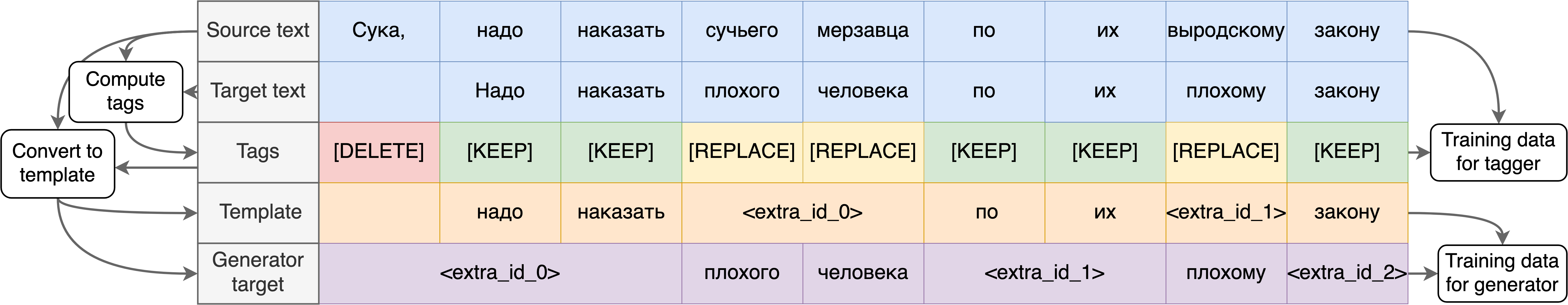}
  \caption{Data generation for training a tagger and a generator.}
  \label{fig:rdtle_training}
\end{figure}

The most direct way is to fine-tune a sequence-to-sequence model on a parallel corpus. Inputs are templates from the tagger, and outputs are masked words from the target sentence. The whole process of data generation for training is in Figure~\ref{fig:rdtle_training}. We also concatenate a source sentence with the generated template, as in \newcite{lewis}, to provide access to the original masked words.

\subsection{Final model}

Our final model uses tagger and generator, both based on a parallel corpus, so we will call it LEWIP (\textbf{L}evenshtein \textbf{e}diting \textbf{wi}th \textbf{p}arallel corpus), following the LEWIS~\cite{lewis} scheme, as there is no <<unsupervised synthesis>> step.

We use conversational RuBERT as a base model for tagger and two versions of the ruT5-base model for generator\footnote{\url{https://huggingface.co/cointegrated/rut5-base}}\footnote{\url{https://huggingface.co/sberbank-ai/ruT5-base}}, with the final submission based on the Sber model. We did not use ruT5-large in a shared task submission, as it did not fit into our GPU memory.

Organizers of the shared task provided a parallel corpus of 11090 pairs for training, a development set of 800 samples, and a test set of 875 samples. We used only that data for the style transfer model.

\section{Results}

\subsection{In-house annotation and automatic metrics}

\begin{table}[t]
\begin{center}
\begin{tabular}{|c|c|c|c|}
\hline \bf Architecture & \bf Generator & \bf STA, \% & \bf SIM, \% \\ \hline
Seq2seq & T5 baseline & 83.5 & \textbf{87.0} \\
Seq2seq & cointegrated/ruT5-base & 73.0 & 83.5 \\
Seq2seq & sber/ruT5-base & 81.0 & 85.0 \\
LEWIP & cointegrated/ruT5-base & 93.5 & 76.0 \\
LEWIP & sber/ruT5-base & \textbf{94.5} & 81.5 \\

\hline
\end{tabular}
\end{center}
\caption{In-house manual annotation with crowdsourcing, 200 random samples from the test set}
\label{tab:in_house} 
\end{table}

Some results of in-house annotation are in Table~\ref{tab:in_house}. In addition to the T5 baseline provided by the organizers, we fine-tuned two ruT5-base models on our own. LEWIP with the Sber generator shows the best style transfer accuracy among all tested models but a lower similarity score than the baseline. Sber ruT5-base is better than the original ruT5-base for this task in both setups.

Examples where LEWIP failed to generate non-toxic texts according to in-house annotation:
\selectlanguage{russian}
\begin{enumerate}
    \item {
        \textbf{Original}: сколько же е**нутых в россии в месте с тобой\\
        Translation: how many f**ked up people in russia are there with you\\
        \textbf{Generated}: сколько же неадекватных в россии в месте с тобой\\
        Translation: how many inadequate people in russia are there with you
    }
    \item {
        \textbf{Original}: какие же эти люди сволочи!!!\\
        Translation: what kind of bastards are these people!!!\\
        \textbf{Generated}: какие же эти люди плохие !\\
        Translation: how bad are these people!
    }
\end{enumerate}

Examples where LEWIP failed to generate similar texts according to in-house annotation:
\selectlanguage{russian}
\begin{enumerate}
    \item {
        \textbf{Original}: у курки е**ные  не оболщатесь это миф.\\
        Translation: f**king s toners, don't be fooled, it's a myth. \\
        \textbf{Generated}: у вас не оболденьтесь это миф\\
        Translation: you don't go crazy, it's a myth.
    }
    \item {
        \textbf{Original}: Только хотел спросить, что за завалы. Е**ть хреновые в Рашке плотники\\
        Translation: Just wanted to ask what are these obstructions. The carpenters in Russia are f**king bad\\
        \textbf{Generated}: Только хотел спросить, что за завалы. в Рашке плотники\\
        Translation: Just wanted to ask what are these obstructions. The carpenters in Russia
    }
\end{enumerate}
\selectlanguage{british}

It seems that the tagger works well in most cases, and problems are mostly in the generator.

\begin{table}[t]
\begin{center}
\begin{tabular}{|c|c|c|c|c|c|c|}
\hline \bf Architecture & \bf Model & \bf Our STA, \% & \bf SIM, \% & \bf FL, \% & \bf J, \% \\ \hline
Seq2seq & T5 baseline & 86.3 & 82.7 & 83.7 & 59.3\\
Seq2seq & cointegrated/ruT5-base & 78.8 & \textbf{85.0} & 83.9 & 55.2 \\
Seq2seq & sber/ruT5-base & 83.8 & 83.6 & 83.4 & 57.8\\
LEWIP & cointegrated/ruT5-base & \textbf{93.6} & 79.7 & 88.4 & \textbf{66.1}  \\
LEWIP & sber/ruT5-base & 93.1 & 79.8 & \textbf{88.5} & 65.8 \\

\hline
\end{tabular}
\end{center}
\caption{Automatic metrics on the test set}
\label{tab:auto} 
\end{table}

Automatic metrics for the same set of models are in Table~\ref{tab:auto}. Joint and STA scores for both LEWIP models are higher than the baseline.

\subsection{Final human evaluation}

\begin{table}[t]
\begin{center}
\begin{tabular}{|c|c|c|c|c|c|}
\hline \bf Team & \bf STA, \% & \bf SIM, \% & \bf FL, \% & \bf J, \% \\ \hline
Human References & 88.8 & 82.4  & 89.4  & 65.3 \\
\hline
T5 baseline  & 79.1 & 82.2 & \textbf{92.5} & 60.6 \\
\hline
SomethingAwful & 79.4 & \textbf{87.2} & 90.3 & \textbf{63.3} \\
FRC CSC RAS & 73.4 & 86.5 & 91.8 & 59.8 \\
Our system & \textbf{82.4} & 79.1 & 84.6 & 58.2 \\
\hline
\end{tabular}
\end{center}
\caption{Final results of the shared task, human evaluation, 3 top teams out of 10}
\label{tab:final_human_eval} 
\end{table}

The final results are in Table~\ref{tab:final_human_eval}. Our model's style transfer accuracy is much worse than our in-house annotation. We explain it with different instructions and annotation protocols. Still, our model has the best style transfer accuracy among all other models but with lower semantic similarity than the baseline. \newcite{pang} showed that these metrics are complementary and challenging to optimize simultaneously.

We attempted to rank several beam search hypotheses from generators with automatic metrics to find different trade-offs, and we were successful in the sense of these automatic metrics. Nevertheless, it did not yield better human assessments. Generators were coming up with adversarial examples that were wrong for humans but good for automatic metrics.

\subsection{Computational effectiveness}

In some cases, we do not need the second step of the system. For 238 examples of 875 (27\%) in the test set, the generator model was not run because there were no <<replace>> or <<insert>> tags. Sequence-to-sequence models are much more computationally expensive than encoder-only taggers. Moreover, a generator requires fewer steps than a raw sequence-to-sequence model, as it only fills the gaps. Overall, our system is more computationally effective than a T5 baseline.

\section{Conclusions}

\begin{itemize}
    \item Thorough testing of a classification model helps in building data augmentations and, eventually, a much more stable model.
    \item Current automatic metrics are not reliable for evaluating systems trained on parallel corpora. They can work in a range of low values, e.g., for unsupervised style transfer, but there are too unstable to work with accurate models.
    \item Text editing models can perform at least as well as pure sequence-to-sequence models. They have the inductive bias based on the assumption that input and output are very close. They are also more environmental-friendly than pure sequence-to-sequence models.
\end{itemize}

\bibliography{dialogue.bib}
\bibliographystyle{dialogue}



\end{document}